\documentclass{article}
\PassOptionsToPackage{numbers, compress}{natbib}

\usepackage[final]{nips_2016}
\usepackage{times}
\usepackage{epsfig}
\usepackage{graphicx}
\usepackage{amsmath}
\usepackage{amssymb}
\usepackage{subcaption}
\usepackage[export]{adjustbox}
\usepackage{parskip}
\usepackage[utf8]{inputenc} % allow utf-8 input
\usepackage[T1]{fontenc}    % use 8-bit T1 fonts
\usepackage{hyperref}       % hyperlinks
\usepackage{url}            % simple URL typesetting
\usepackage{booktabs}       % professional-quality tables
\usepackage{amsfonts}       % blackboard math symbols
\usepackage{nicefrac}       % compact symbols for 1/2, etc.
\usepackage{microtype}      % microtypography

\begin{document}

%%%%%%%%% TITLE
\title{Breast Mass Classification from Mammograms using Deep Convolutional Neural Networks}

\author{Daniel Lévy, Arzav Jain\\
Stanford University\\
\texttt{\{danilevy,ajain\}@cs.stanford.edu}}

\maketitle
%\thispagestyle{empty}

%%%%%%%%% ABSTRACT
\begin{abstract}
Mammography is the most widely used method to screen breast cancer. Because of its mostly manual nature, variability in mass appearance, and low signal-to-noise ratio, a significant number of breast masses are missed or misdiagnosed. In this work, we present how Convolutional Neural Networks can be used to directly classify pre-segmented breast masses in mammograms as benign or malignant, using a combination of transfer learning, careful pre-processing and data augmentation to overcome limited training data. We achieve state-of-the-art results on the DDSM dataset, surpassing human performance, and show interpretability of our model. 
\end{abstract}

%%%%%%%%% BODY TEXT
\section{Introduction}
According to the International Agency for Research on Cancer, breast cancer accounts for $22.9\%$ of invasive cancers and $13.7\%$ of cancer-related deaths in women worldwide \citep{boyle2008world}. In the U.S., $1$ in $8$ women is expected to develop invasive breast cancer over the course of her lifetime \cite{desantis2014breast}. Routine mammography is standard for preventive care and detection of breast cancer before biopsy. However, it is still a manual process, prone to human error due to high variability in mass appearance \cite{ball2007digital} and low signal-to-noise ratio, and thus can cause unnecessary biopsies or worse, missed malignant masses. Furthermore, efficacy is often highly correlated with radiologist expertise and workload \cite{elmore2009variability}.

Convolutional Neural Networks (CNN) have achieved impressive results on computer visions tasks spanning classification \cite{Krizhevsky2012}, object detection \cite{girshick2014rich}, and segmentation \cite{long2015fully}. For breast mass diagnosis, deep learning techniques have been explored \cite{arevalo2016representation, carneiro2015unregistered, dhungel2015automated, dhungel2015deep, wang2016discrimination}, but never in a fully end-to-end manner (directly classifying from pixel space) because of the scarcity of available data and lack of interpretability. In this work, we successfully train end-to-end CNN architectures to directly classify breast masses as benign or malignant. We obtain state-of-the-art results using a combination of transfer learning, careful pre-processing and data augmentation. We analyze the effects of these modelling choices and furthermore show how we can provide interpretability to the model's predictions.

\section{Related Work}

While medical images differs significantly from natural images, traditional feature engineering techniques from computer vision such as scale-invariant feature transform (SIFT) and histogram of oriented gradients (HOG) have seen use and success when applied to medical images. More recently, deep learning-based approaches using CNNs have begun to achieve impressive performance on medical tasks such as chest pathology identification in X-Ray and CT~\cite{bar2015chest, van2015off}, and thoraco-abdominal lymph node detection and interstitial lung disease classification \cite{shin2016deep}. 

In the context of mammography, \cite{dhungel2015automated, dhungel2015deep} detect breast masses using a combination of R-CNN and random forests. Multiple works tackle the problem of breast lesion classification, but typically adopt a multi-stage approach. \cite{wang2016discrimination} extracts hand-engineered semantic (such as calcification) and textual features, and \cite{carneiro2015unregistered} classifies a full mammogram by extracting features from each view of the breast and combining them to output a prediction. \cite{arevalo2016representation} performs extensive pre-processing using domain knowledge before training a CNN. To the best of our knowledge, ours is the first work to directly classify pre-detected breast masses using CNN architectures.

\section{Dataset}

\begin{figure}[h!]
\centering
\begin{subfigure}{0.48\textwidth}
\centering
\includegraphics[height=1.5cm,width=1.5cm]{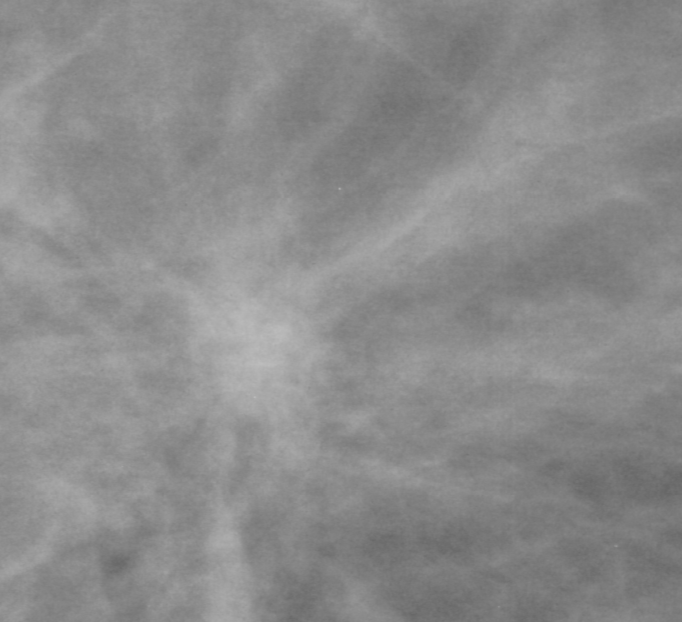}
\includegraphics[height=1.5cm,width=1.5cm]{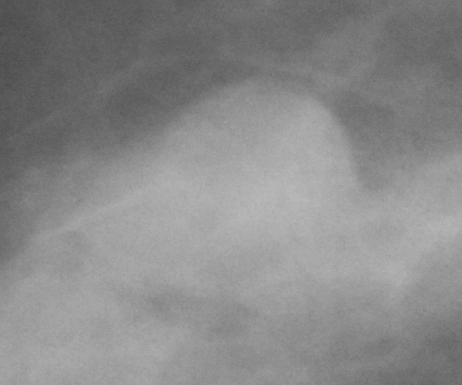}
\includegraphics[height=1.5cm,width=1.5cm]{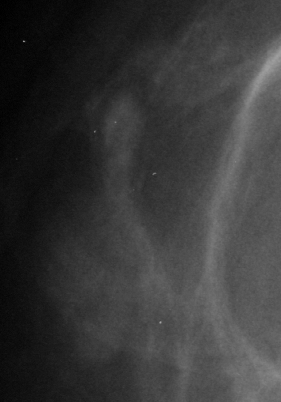}
\includegraphics[height=1.5cm,width=1.5cm]{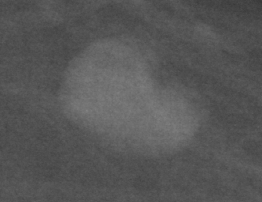}
%\caption{Benign}
\label{fig:TopFeatures}
\end{subfigure}
\begin{subfigure}{0.48\textwidth}
\centering
\includegraphics[height=1.5cm,width=1.5cm]{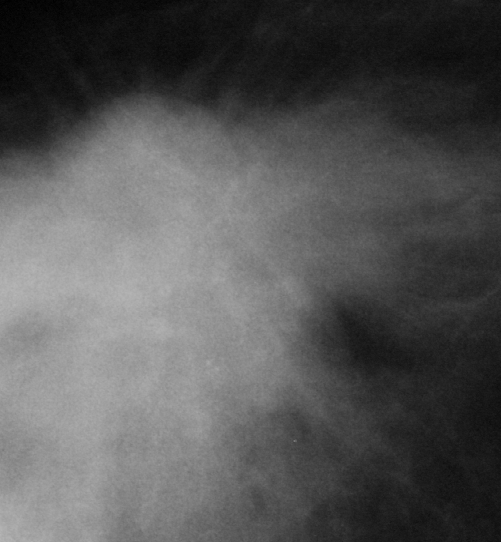}
\includegraphics[height=1.5cm,width=1.5cm]{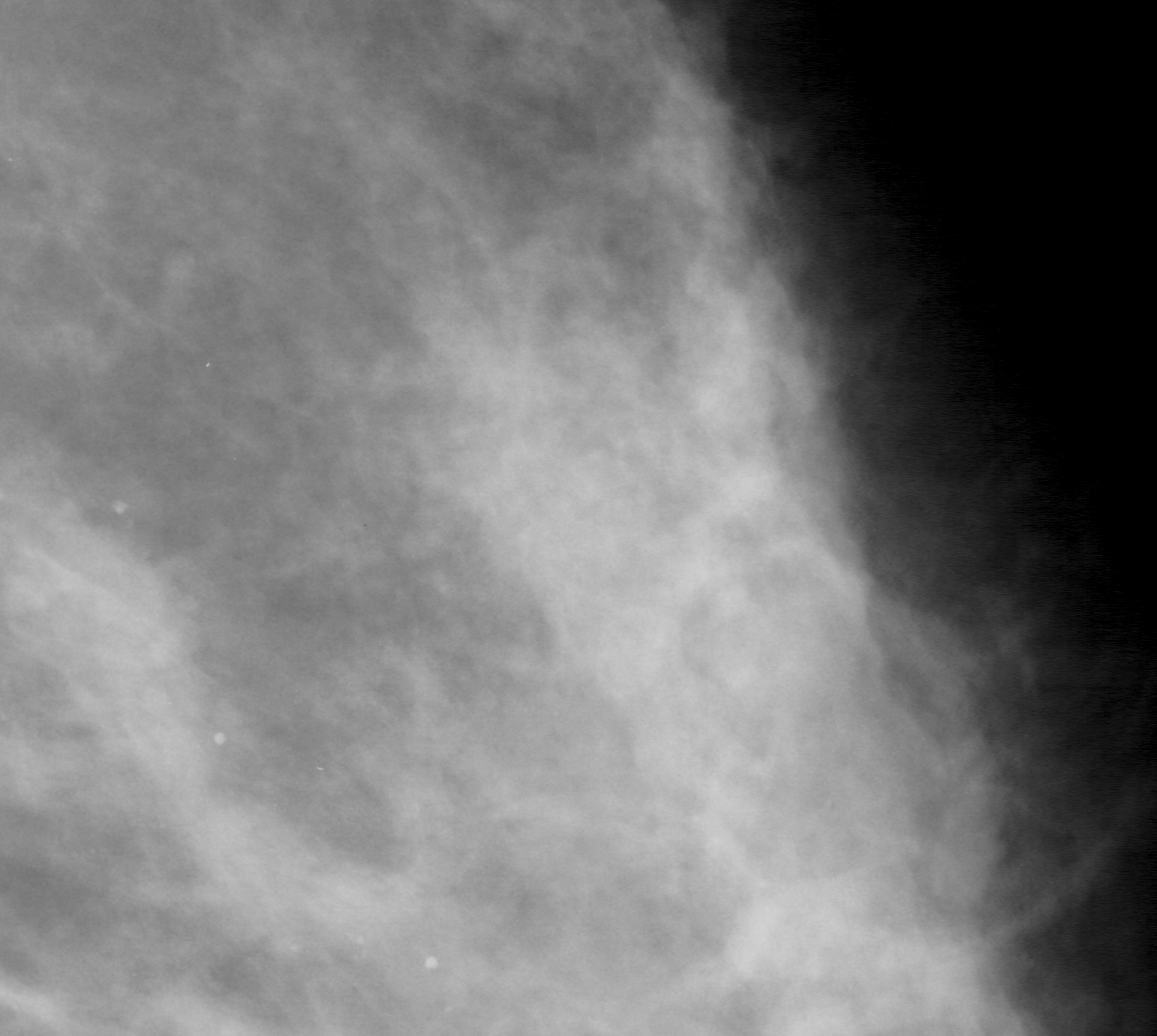}
\includegraphics[height=1.5cm,width=1.5cm]{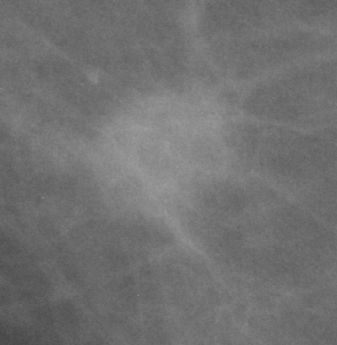}
\includegraphics[height=1.5cm,width=1.5cm]{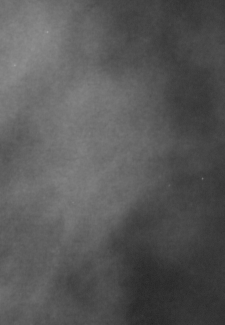}
%\caption{Malignant}
\label{fig:FeatureSets}
\end{subfigure}
\caption{Benign (Left) and malignant (Right) breast masses from the dataset.}
\label{fig:sampleInput}
\end{figure}

In our experiments, we use the Digital Database for
Screening Mammography (DDSM) \cite{bowyer1996digital}, a collaboratively maintained public dataset at the University of South Florida. It comprises approximately 2500 studies each containing both mediolateral oblique (MLO) and craniocaudal (CC) views of each breast. Each image is grayscale and accompanied by a mask specifying the region of the pre-segmented mass if present. Examples of benign and malignant masses are shown in Fig.~\ref{fig:sampleInput}.

We consider only mammograms which contain masses, resulting in 1820 images from 997 patients. We split these randomly by patient into training, validation and testing sets ($80\%, 10\%$ and $10\%$ of the full dataset), constraining the validation and test sets to be balanced.

\section{Methods}
\label{methods}

We train three different CNN architectures for breast mass classification, and analyze the effect of a number of model choices. We describe these below.

\subsection{CNN architectures}

We evaluate three network architectures: a shallow CNN (the baseline model), an AlexNet \cite{Krizhevsky2012} and a GoogLeNet \cite{szegedy2015going}. For both the AlexNet and GoogLeNet, we use the same base architecture as the original works but replace the last fully-connected (FC) layer to output $2$ classes. We also remove the two auxiliary classifiers from the GoogLeNet which we found impaired our training in practice.

The baseline model's architecture is inspired by the early layers of AlexNet \cite{Krizhevsky2012}. We additionally use batch normalization \cite{ioffe2015batch}. The network takes a $224 \times 224 \times 3$ image as input. It consists of $3$ convolutional blocks composed of $3 \times 3$ Convolutions - Batch Norm - ReLU - Max Pooling, with respectively $32$, $32$ and $64$ filters each, followed by $3$ FC layers of size $128$, $64$ and $2$. The final layer is a soft-max layer for binary classification. We use ReLU activation functions, Xavier~\cite{glorot2010understanding} weight initialization, and the Adam~\cite{kingma2014adam} update rule with a base learning rate of $10^{-3}$ and batch size $64$. 

\subsection{Aspects of model analysis}
\label{sec:aspects}

\paragraph{Transfer learning} We analyze the effect of initializing networks with pre-training on the ImageNet dataset~\cite{russakovsky2015imagenet}, and then fine-tuning on mammography images. For the AlexNet model, we initialize the convolutional layers with pre-trained weights and a smaller learning rate multiplier of $0.1$, and randomly initialize the 3 FC layers. For the GoogLeNet, we use the same weight initialization scheme. We use a learning rate multiplier of $0.1$ for the layers before the Inception\_5a module, $1$ for the Inception\_5a and Inception\_5b modules, and $10$ for the last FC layer for more aggressive learning.

We train the AlexNet with Adam, base learning rate $10^{-3}$, and dropout rate $0.5$. We train the GoogLeNet with Vanilla SGD, base learning rate $10^{-2}$, and dropout rate $0.2$.

%\paragraph{Context} We apply standard per-processing: we extract the mass from the full mammogram by taking a bounding box around the pixel-level mask, we subtract the mean pixel computed over the whole training set and replicate the grayscale values across the 3 channels to fit the RGB input from the pre-trained networks.

\paragraph{Mass context} The area surrounding a mass provides useful context for diagnosis. We explore two approaches for providing the network with context. In the first, our input to the network is the region including $50$ pixels of fixed padding around the mass, providing a context size independent of mass dimensions (referred to as \textbf{Small Context}). In the second approach we use proportional padding by extracting a region two times the size of the mass bounding box (referred to as \textbf{Large Context}).

\paragraph{Data Augmentation} We study the impact of data augmentation to alleviate to the relatively small size of our training set, which is characteristic of many medical image datasets. We use rotation, cropping, and mirroring transformations to increase the effective size of our dataset (referred to as \textbf{Aug}). For each training image, we perform 5 random rotations and sample 5 random crops per rotation offline, effectively increasing training set size by a factor of 25. We also perform random mirroring at train time. These augmentations are justified since masses have no inherent orientation and their diagnosis is invariant to these transformations.

\section{Results}

We first present empirical analysis of our model design using the AlexNet base architecture, then show quantitative results of our best models. Finally we use techniques for visualizing saliency maps to provide interpretability of the model. All experiments are implemented with Caffe~\cite{jia2014caffe}, and the analysis and results are presented on the validation and test sets, respectively.

\subsection{Empirical analysis}
\label{sec:empirical_analysis}

\paragraph{Effectiveness of transfer learning} CNN-based image representations learned on large-scale annotated datasets have proven to to be a useful form of pre-training that can be effectively transferred to other computer vision tasks with limited training data \cite{oquab2014learning}. More recently, low-level features learned from natural images have shown to be effective for medical image classification \cite{bar2015chest, shin2016deep, van2015off}. In Table \ref{table:transfer}, we strongly confirm this claim by demonstrating that a fine-tuned AlexNet significantly outperforms our baseline model.

\begin{table}[h!]
\begin{center}
\begin{tabular}{|l|c|}
\hline
Model & Validation Accuracy \\
\hline\hline
Baseline(Aug-Large Context) & $0.66$ \\
AlexNet(Aug-Large Context)  & $\mathbf{0.90}$ \\
\hline
\end{tabular}
\end{center}
\caption{Effectiveness of transfer learning.}
\label{table:transfer}
\end{table}

\paragraph{Influence of context} To understand the influence of context around masses, we fine-tune an AlexNet on two different datasets - one with fixed padding and the other with proportional padding. The results in Table \ref{table:padding} show that proportional padding contains greater signal for classification, and we consequently use this for the remaining experiments.

\begin{table}[h!]
\begin{center}
\begin{tabular}{|l|c|}
\hline
Model & Validation Accuracy \\
\hline\hline
AlexNet(No Aug-Small Context) & $0.64$ \\
AlexNet(No Aug-Large Context)  & $\mathbf{0.71}$ \\
\hline
\end{tabular}
\end{center}
\caption{Influence of context around the breast mass on the model performance.}
\label{table:padding}
\end{table}

\begin{figure}[h!]
\centering
\includegraphics[scale=1.0]{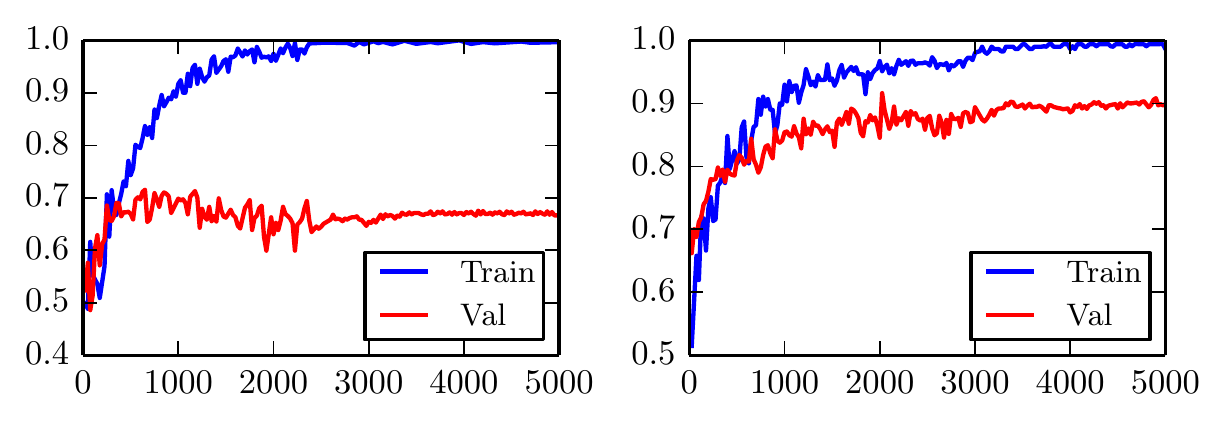}
\caption{Accuracy curves when training on the unaugmented (Left) vs.  augmented dataset (Right). x-axis is the iteration number.}
\label{fig:dataAug}
\end{figure}

\paragraph{Influence of data augmentation} Limited amounts of training data is a common bottleneck in machine learning applications to medical problems. We evaluate the utility of data augmentation schemes to increase the effective amount of training data and reduce overfitting. The training loss curves in Fig.~\ref{fig:dataAug} show that our data augmentation technique described in Sec.~\ref{sec:aspects} successfully regularises the network and helps remedy the scarcity of data.

%\subsection{GoogLeNet: Deeper vs Shallow Training}
%As discussed in section \ref{section:GoogLeNetmodel}, two GoogLeNet models were fine-tuned with different parameters. The Shallow model converged within 20 epochs while the Deeper model converged withing 40 epochs. This makes sense since the Deeper model requires more iterations to better learn the high-level features at both the \textit{inception\_5a} and \textit{inception\_5b} modules. 
%
%Four snapshots of each model were also taken at regular intervals during fine-tuning. We show the performance of each of these snapshots on the validation set in Figure \ref{fig:snapshot-pr}. Although the accuracy of the two models closely follow each other, the Deeper model does better on recall throughout whereas the Shallow model does better on precision. Note that the accuracy, precision and recall for the Deeper model is best at the $0.75$ mark. Thus, we take this snapshot at $30$ epochs ($= 0.75 \times 40$) to be our final Deeper model in the rest of this paper (and in particular, Table \ref{table:summary}).
%
%\begin{figure}[h!]
%\centering
%\includegraphics[width=8cm]{deepershallow.png}
%\caption{Validation accuracy, precision and recall for the Deeper Training and Shallow Training GoogLeNet models. The x-axis represents the fraction of the total number of epochs which is $40$ for the Deeper model and $20$ for the Shallow model.}
%\label{fig:snapshot-pr}
%\end{figure}
\subsection{Performance}

Our final results using the model choices described in Sec.~\ref{sec:empirical_analysis} and all base architectures are presented in Table \ref{table:summary}. The GoogLeNet outperforms the other models by a fair margin. It is also more suited for fine-tuning and less prone to overfitting due to its relatively small number of parameters, approximately $5$ million compared to $100$ million for AlexNet. 

An important metric for diagnostic applications is maximizing recall, since the cost of a false negative (patient remaining undiagnosed) is much higer than a false positive (an additional biopsy). Our best model achieves $0.934$ recall at $0.924$ precision, outperforming human performance in a study that shows radiologist recall between $0.745$ and $0.923$~\cite{elmore2009variability}. This result is very promising for real-life use of such models in clinical practice.

\begin{table*}[ht!]
\centering
\begin{tabular}{|l || c | c | c | c |} 
\hline
Model & Accuracy & Precision & Recall & \# Epochs \\
\hline\hline
Baseline (Aug-Large Context) & $0.604$ & $0.587$ & $0.703$ & $35$\\
AlexNet (Aug - Large Context) & $0.890$ & $0.908$ & $0.868$ & $30$ \\
%GoogLeNet (Aug - Large Context) - Shallow Training & 0.912 & 0.921 & 0.901 & 20 \\
GoogLeNet (Aug - Large Context) & $\mathbf{0.929}$ & $\mathbf{0.924}$ & $\mathbf{0.934}$ & $30$ \\
\hline
\end{tabular}
\caption{Summary of performance on the test set.}
\label{table:summary}
\end{table*}

\subsection{Interpretability}

\begin{figure}[h!]
\begin{subfigure}{0.5\textwidth}
\centering
\includegraphics[width=6cm]{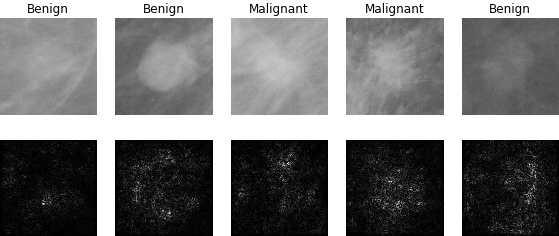}
\caption{AlexNet}
\end{subfigure}
\begin{subfigure}{0.5\textwidth}
\centering
\includegraphics[width=6cm]{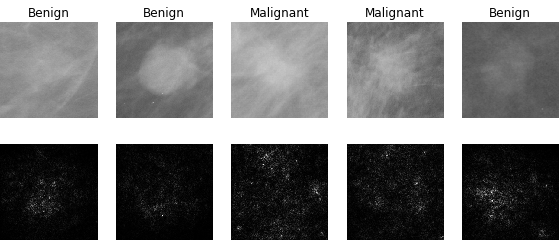}
\caption{GoogLeNet}
\end{subfigure}
\caption{Saliency maps for our best AlexNet and GoogLeNet on five images from the validation set.}
\label{fig:saliency}
\end{figure}

Deep learning models often lack interpretability and as such are hard to adopt for practical use in medical settings. \cite{simonyan2013deep} describe a methodology to visualize saliency maps which show the regions of an image the network is sensitive to when making predictions. This is performed by computing the gradient of the image with respect to the unnormalized class scores. Regions with larger gradient indicate higher contribution to the prediction (brighter in Fig.~\ref{fig:saliency}). Both the AlexNet and GoogLeNet learn to attend to the edges of the mass, which is a high-signal criterion for diagnosis, while also paying attention to context.

%\subsection{Conservativeness}
%
%Shown in Table \ref{table:conservative} are the accuracies and recall of three models on the validation set. We see that recall is usually in the range of or greater than precision, thus suggesting that our models correctly classify a larger fraction of the malignant masses (fewer false negatives) than benign masses. This conservative property of our models is arguably desired given that we don't want to misdiagnose malignant masses.
%
%\begin{table}[h!]
%\begin{center}
%\begin{tabular}{|l|c|c|}
%\hline
%Model & Precision &  Recall\\
%\hline\hline
%LevyNet & 0.649 & $\mathbf{0.692}$ \\
%AlexNet (Aug - Large Context) & 0.904 & $\mathbf{0.934}$ \\
%GoogLeNet-Shallow Training & 0.920 & 0.890 \\
%GoogLeNet-Deeper Training & 0.912 & $\mathbf{0.923}$ \\
%\hline
%\end{tabular}
%\end{center}
%\caption{Model metrics on the validation set.}
%\label{table:conservative}
%\end{table}

\section{Conclusion}

In this work, we propose an end-to-end deep learning model to classify pre-detected breast masses from mammograms. We show how careful pre-processing, data augmentation and transfer learning can overcome the data bottleneck common to medical computer vision tasks, and additionally provide a method to give more interpretability to network predictions. 

Our approach obtains state-of-the-art results, outperforming trained radiologists, and the interpretability enables more comfortable adoption in real-world settings. Future work includes exploring other architectures, and integration of attention mechanisms which are more difficult to train but could provide even more concrete interpretability.

\subsection*{Acknowledgements}

We thank Justin Johnson for continuous feedback and guidance throughout this project as well as Serena Yeung for insightful comments on the draft. The authors also ackowledge the support of AWS Educate program for generously providing free instances with GPUs.

\bibliography{refs}
\bibliographystyle{abbrvnat}

\end{document}